\documentclass{article}

\PassOptionsToPackage{numbers,sort&compress}{natbib}
\usepackage{graphicx}
\usepackage{multirow}
\usepackage[ruled,vlined]{algorithm2e}
\SetKwInOut{Input}{Input}
\SetKwInOut{Output}{Output}
\usepackage{amsmath}
\usepackage{placeins}
\usepackage[table]{xcolor}
\usepackage{xcolor}
\usepackage{enumitem}

 \usepackage[preprint]{format}

\usepackage[utf8]{inputenc} 
\usepackage[T1]{fontenc}    
\usepackage{hyperref}       
\usepackage{url}            
\usepackage{booktabs}       
\usepackage{amsfonts}       
\usepackage{nicefrac}       
\usepackage{microtype}      
\usepackage{xcolor}         

\title{Context without Commitment: Robust Dense Correspondence under Non-Rigid Deformation}

\author{%
  Yuzhen He \\
  Mannheim Institute for Intelligent Systems in Medicine (MIISM)\\
  Medical Faculty Mannheim, Heidelberg University, Mannheim, Germany\\
  \And
  Sara Homscheid\thanks{Corresponding author: \texttt{sara.homscheid@medma.uni-heidelberg.de}} \\
  Mannheim Institute for Intelligent Systems in Medicine (MIISM)\\
  Medical Faculty Mannheim, Heidelberg University, Mannheim, Germany\\
  }

\begin{document}
\maketitle

\begin{abstract}
Non-rigid point-cloud registration aims to find the corresponding target point for every point on a deforming source surface. Point-level matching keeps the complete target cloud available, but correspondence can become ambiguous when different regions have similar local geometry. Regional or coarse-to-fine methods provide larger spatial context, but an incorrect regional match can exclude the correct point correspondence before the final dense matching stage. We propose CoCo-Reg, which uses regional patches to enrich dense point features without allowing patch predictions to restrict the final point-level search. CoCo-Reg constructs farthest-point-sampled patches, exchanges geometric information within and between the source and target, supervises patch similarity using identity-corrected point overlap, and projects the resulting regional information back to the dense point features. The final registration stage still scores the complete target cloud before performing its global point-level candidate selection. On 726 held-out ModelNet10 objects across nine deformation levels, two established learning-based non-rigid registration baselines obtain mean correspondence errors of 0.1993 and 0.1921, whereas CoCo-Reg obtains 0.0547. Relative to the point-level baseline on which CoCo-Reg is built, this corresponds to a 72.6\% reduction. CoCo-Reg achieves lower correspondence error on 92.3\% of paired test objects and reduces the mean fraction of points with error above 0.1 from 47.3\% to 17.3\%. Chamfer distance and HD95 decrease in the same direction, and CoCo-Reg remains lower across all tested deformation levels. These results support the use of regional context for dense non-rigid correspondence without imposing a hard patch-level restriction on the final search. Because the evaluation uses one checkpoint per method, the reported gains characterize the complete evaluated systems rather than the isolated causal contribution of an individual component. Code will be made publicly available.
\end{abstract}

\section{Introduction and Related Work}
\label{sec:intro}
Non-rigid point-cloud registration estimates how a source surface moves to a target surface by predicting a spatially varying displacement for its points. Unlike rigid registration, one global transformation is not sufficient: different parts of a deformable object may move in different directions and by different amounts. Reliable correspondence is therefore difficult when deformation is large, local neighborhoods change, or several parts of the shape look geometrically similar. This problem appears in deformable-object analysis, 3D scene understanding, and soft-tissue modeling, where tissue shift can invalidate a fixed preoperative geometry \citep{monjiazad2023review,maennle2023digitaltwin,monjiazad2024digitaltwin,robustdefreg2025,deftransnet2025}.

\textbf{\textit{Point-level learned matching.}} Many learning-based registration methods keep correspondence reasoning at the point level. DCP, RPM-Net, PREDATOR, PointDSC, and RegTR learn point descriptors and correspondence rules for rigid or partially overlapping point clouds \citep{wang2019dcp,yew2020rpmnet,huang2021predator,bai2021pointdsc,yew2022regtr}. For deformable geometry, FLOT learns dense motion through transport, Lepard introduces cross-cloud positional information, and Neural Deformation Pyramid models non-rigid motion hierarchically \citep{puy2020flot,li2022lepard,li2022ndp}. Recent non-rigid methods such as DefTransNet keep a global point-level candidate search, while DINE adds a prior over the complete deformation field \citep{deftransnet2025,dine2026}. Point-level matching is flexible because every target point can remain a candidate, but the descriptor of one point or a small neighborhood may be insufficient to distinguish repeated or symmetric regions under strong deformation.

\textbf{\textit{Coarse-to-fine and regional matching.}} A second family first reasons over larger regions and then performs finer matching. PointNet++ and DGCNN established hierarchical and neighborhood-based point representations, while CoFiNet and GeoTransformer use superpoints or coarse correspondences to guide registration \citep{qi2017pointnetpp,wang2019dgcnn,yu2021cofinet,qin2023geotransformer}. Robust-DefReg similarly uses a coarse-to-fine strategy for non-rigid registration \citep{robustdefreg2025}. Regional reasoning can reduce local ambiguity, but it introduces another failure mode: if a coarse source region is matched to the wrong target region and that decision is used to restrict the fine search, the correct point correspondence may no longer be reachable. This is particularly important for non-rigid deformation because source and target patches are constructed independently and one physical region can overlap several patches after deformation.

\textbf{\textit{Patch representations.}} Patch-based representation learning provides a useful middle ground between individual points and whole objects. Point-BERT and Point-MAE use local patches for masked representation learning, and PointGPT uses ordered point patches for autoregressive pretraining \citep{yu2022pointbert,pang2022pointmae,chen2023pointgpt}. PointGPT motivated our use of structured patch tokens, but CoCo-Reg does not use its pretrained model, tokenizer, causal ordering, or autoregressive objective. Instead, patches are constructed directly from the registration clouds and are used only to provide regional information for dense correspondence.

The remaining challenge is therefore not simply to choose between point-level and patch-level matching. Point-level search preserves flexibility but may lack enough context; hard patch-based matching provides context but can make an early mistake irreversible. CoCo-Reg addresses this trade-off by using patches to \emph{inform} the dense point features without using patch predictions to define the final search space. Each cloud is partitioned into farthest-point-sampled patches, patch tokens exchange geometry-aware information within each cloud and across the source and target, and patch similarity is trained from identity-corrected point overlap. The contextual patch token is then projected back to the points of that patch as a residual feature update. The original dense registration stage still ranks candidates over the complete target cloud. We refer to this simple design as \emph{context without commitment}: patches provide guidance, but they do not decide which target region a point is allowed to search. Our main contribution is therefore not a new dense registration backbone, but a way to add regional information without turning the regional prediction into hard candidate pruning.

Our contributions are:
\begin{itemize}[leftmargin=*,nosep]
    \item We introduce a patch-context extension for dense non-rigid registration that adds larger-scale regional information to point features while preserving the original global point-level correspondence search.
    \item We formulate identity-corrected, multi-positive patch supervision for independently constructed source and target patches, allowing one source patch to overlap more than one valid target patch after deformation.
    \item We evaluate the complete CoCo-Reg configuration on 726 paired held-out objects across deformation levels 0.1--0.9 using correspondence-specific, set-level, tail-error, failure-rate, and paired statistical analyses, and explicitly separate the observed system-level gains from stronger component-level causal claims.
\end{itemize}

\section{Proposed Method}
\label{sec:method}
\textbf{\textit{Problem definition.}} Let the source cloud be $S=\{s_i\}_{i=1}^{N}\subset\mathbb{R}^3$ and let the identity-indexed, unshuffled target be $T^{\star}=\{t_i\}_{i=1}^{N}$. During training and evaluation the target tensor is shuffled, so the network receives $T=\{t_{\pi(r)}\}_{r=1}^{N}$ for a permutation $\pi$. The correspondence identity is not given to the network; it is retained only for supervision and evaluation. The ground-truth displacement of source point $i$ is
\begin{equation}
 d_i^{\star}=t_i-s_i,
 \qquad
 \hat D=\{\hat d_i\}_{i=1}^{N},
\end{equation}
and the primary dense objective is the mean absolute component-wise displacement error,
\begin{equation}
 \mathcal L_{\mathrm{point}}=\frac{1}{3N}\sum_{i=1}^{N}\|\hat d_i-d_i^{\star}\|_1.
\end{equation}
The final implementation uses $N=1024$ points per cloud.

\textbf{\textit{Dense backbone features.}} CoCo-Reg retains the DefTransNet feature and displacement backbone \citep{deftransnet2025}. Source and target are mapped to dense 64-dimensional descriptors
\begin{equation}
 (F^S,F^T)=\Phi_{\mathrm{Def}}(S,T),
 \qquad F^S,F^T\in\mathbb{R}^{N\times64}.
\end{equation}
These same descriptors feed both the patch-context branch and the final dense registration stage.

\textbf{\textit{Patch construction and token initialization.}} For either cloud $P$, farthest-point sampling selects $M=32$ center indices $I=\mathrm{FPS}(P,M)$ with coordinates $C=P[I]$. Every dense point is assigned to its closest center,
\begin{equation}
 g(n)=\arg\min_{m\in\{1,\dots,M\}}\|p_n-c_m\|_2^2,
\end{equation}
forming independent source and target ownership sets. At most $K_p=24$ closest owned points are retained for overlap supervision. Importantly, the token is initialized from the FPS-center feature, not from a pooled patch descriptor. With $H=F[I]\in\mathbb{R}^{M\times64}$,
\begin{equation}
 U=\mathrm{LN}(HW_f+b_f),\qquad
 z_m^{0}=\mathrm{Norm}_2\!\left(u_m+0.1\,\rho(c_m)\right),
\end{equation}
where $W_f$ projects to 128 dimensions and $\rho$ is a two-layer MLP on the center coordinate.

\textbf{\textit{Geometry-aware patch interaction.}} Within each cloud, four-head self-attention is biased by the normalized Euclidean distance between patch centers. For head $r$,
\begin{equation}
 e_{ij}^{(r)}=
 \frac{(q_i^{(r)})^\top k_j^{(r)}}{\sqrt{d_h}}
 +\eta_r\!\left(\frac{\|c_i-c_j\|_2}{\mu}\right),
\end{equation}
where $\mu$ is the median pairwise center distance and $\eta_r$ is a learned MLP. Source and target tokens then exchange information bidirectionally,
\begin{align}
 Z^{S\leftarrow T}&=Z^S+\mathrm{MHA}\!\left(\mathrm{LN}(Z^S),Z^T,Z^T\right),\\
 Z^{T\leftarrow S}&=Z^T+\mathrm{MHA}\!\left(\mathrm{LN}(Z^T),Z^S,Z^S\right),
\end{align}
followed by a residual feed-forward layer with expansion factor two. Patch similarity is a scaled cosine score
\begin{equation}
 a_{ij}=\gamma\,
 \frac{(z_i^S)^\top z_j^T}{\|z_i^S\|_2\|z_j^T\|_2},
\end{equation}
with learned scale $\gamma$, initialized to 8 and constrained to the interval $[1,30]$ during the forward pass.

\textbf{\textit{Identity-corrected patch-overlap supervision.}} Because target tensor positions are shuffled, overlap supervision is computed in point-identity space rather than tensor-position space. Let $R_i^S$ and $R_j^T$ denote the retained source and target tensor indices owned by patches $i$ and $j$, respectively. Since target position $r$ contains point $t_{\pi(r)}$, define the corresponding identity sets as $I_i^S=R_i^S$ and $I_j^T=\{\pi(r):r\in R_j^T\}$. The symmetric overlap is
\begin{equation}
 o_{ij}=\frac{1}{2}\left(
 \frac{|I_i^S\cap I_j^T|}{|I_i^S|}
 +\frac{|I_i^S\cap I_j^T|}{|I_j^T|}
 \right).
\end{equation}
Let $V_T=\{1,\ldots,M\}$ and $V_S^+=\{i:\max_j o_{ij}>0\}$. For each $i\in V_S^+$, pairs with $o_{ij}\ge0.10$ form the positive set $Q_i$; if this set is empty despite non-zero overlap, the maximum-overlap target is used as a fallback. The source-to-target multi-positive loss is
\begin{equation}
 \mathcal L_{\mathrm{MP}}^{S\rightarrow T}
 =-\frac{1}{|V_S^+|}\sum_{i\in V_S^+}
 \log\frac{\sum_{j\in Q_i}\exp(a_{ij})}
 {\sum_{j\in V_T}\exp(a_{ij})},
\end{equation}
and the implementation also applies the analogous target-to-source term. Writing $V_T^+=\{j:\max_i o_{ij}>0\}$, the bidirectional aggregate is
\begin{equation}
 \mathcal L_{\mathrm{MP}}=\tfrac12\!\left(\mathcal L_{\mathrm{MP}}^{S\rightarrow T}+\mathcal L_{\mathrm{MP}}^{T\rightarrow S}\right).
\end{equation}
A complementary best-overlap cross-entropy uses $y_i=\arg\max_j o_{ij}$ and $x_j=\arg\max_i o_{ij}$,
\begin{equation}
\begin{split}
 \mathcal L_{\mathrm{best}}=-\tfrac12\!\Bigg[&\frac{1}{|V_S^+|}\sum_{i\in V_S^+}\log\frac{\exp(a_{i y_i})}{\sum_{j\in V_T}\exp(a_{ij})}\\
 +&\frac{1}{|V_T^+|}\sum_{j\in V_T^+}\log\frac{\exp(a_{x_j j})}{\sum_{i=1}^{M}\exp(a_{ij})}\Bigg].
\end{split}
\end{equation}
Both patch terms use unit weight in the reported configuration. Thus the patch loss is
\begin{equation}
 \mathcal L_{\mathrm{patch}}=\mathcal L_{\mathrm{MP}}+\mathcal L_{\mathrm{best}}.
\label{eq:patch}
\end{equation}

\begin{figure*}[t]
    \centering
    \includegraphics[width=0.9\textwidth]{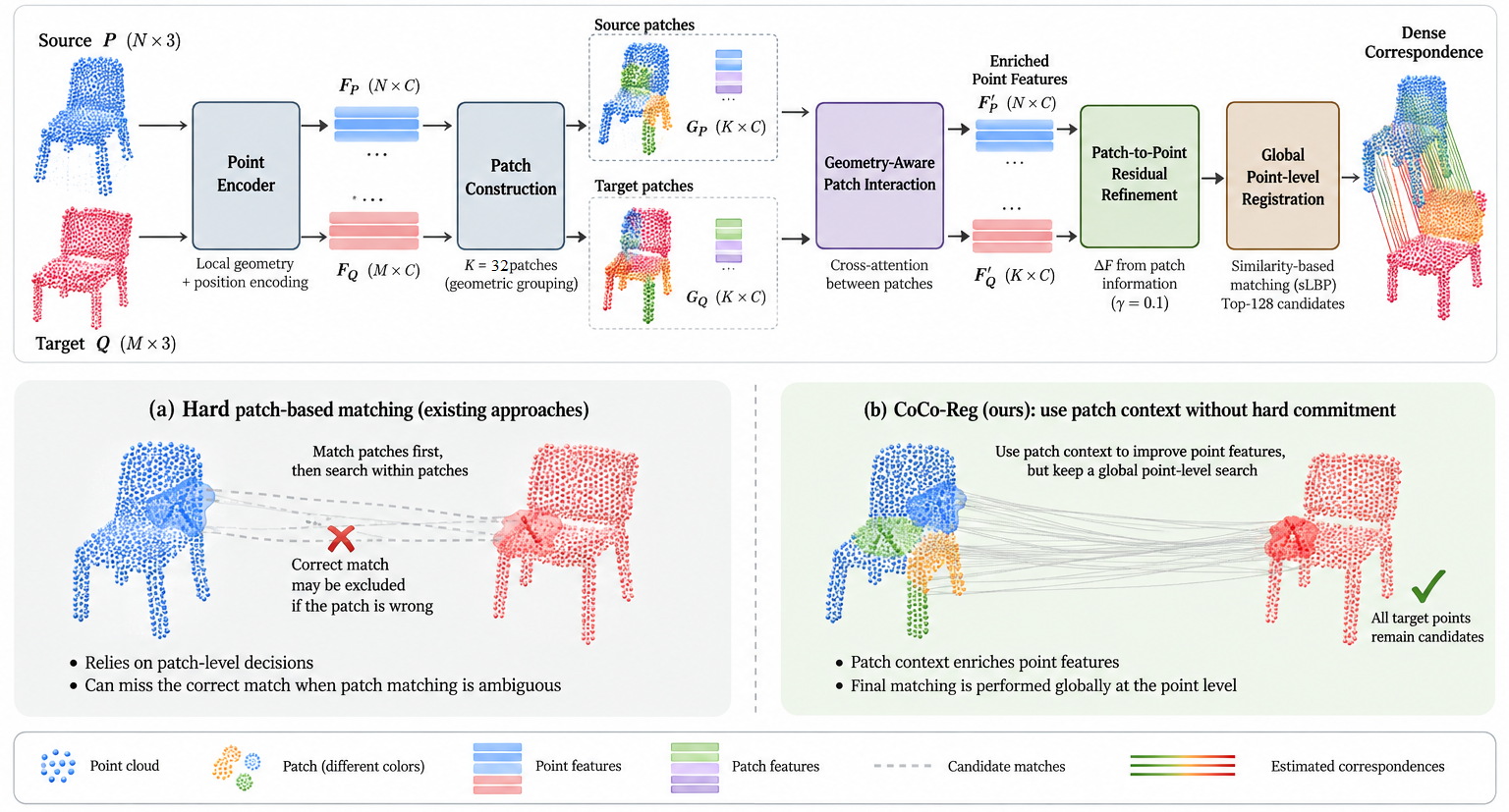}
    \caption{Overview of CoCo-Reg.
    \textbf{Top:} The proposed method extracts dense point features from the source and target point clouds, constructs geometric patches, and exchanges information between source and target patches through geometry-aware interaction. The resulting patch information is projected back to the dense point features through a residual refinement. Final correspondence estimation is then performed at the point level using the global sLBP search, which scores the complete target cloud before retaining the Top-128 candidates.
    \textbf{Bottom:} Comparison with hard patch-based matching. In a hard coarse-to-fine strategy, an incorrect patch match can remove the correct point correspondence from the subsequent search. CoCo-Reg instead uses patch information to enrich the point representation without applying a hard patch-level restriction to the final correspondence search.}
    \label{fig:method}
\end{figure*}

\textbf{\textit{Context back to points without restricting the final search.}} As summarized in Fig.~\ref{fig:method}, each source and target point gathers the final token of its owned patch, $q_n^S=z_{g^S(n)}^S$ and $q_n^T=z_{g^T(n)}^T$, and receives a residual refinement
\begin{equation}
 \widetilde f_n^S=f_n^S+0.10\,\psi(q_n^S),
 \qquad
 \widetilde f_n^T=f_n^T+0.10\,\psi(q_n^T),
\end{equation}
where $\psi:\mathbb{R}^{128}\rightarrow\mathbb{R}^{64}$. The final displacement remains
\begin{equation}
 \hat D=\mathrm{sLBP}(S,T,\widetilde F^S,\widetilde F^T),
 \qquad
 \mathcal L_{\mathrm{total}}=\mathcal L_{\mathrm{point}}+0.10\,\mathcal L_{\mathrm{patch}}.
\label{eq:total}
\end{equation}
For each source point, sLBP scores all target points and only then retains its global Top-128 feature candidates. Thus patch context can change the dense feature ranking, but an incorrect patch relation cannot remove a target point simply because that target lies outside a selected patch. This operational definition is the paper's \emph{context without commitment} principle: the phrase refers specifically to avoiding an early hard patch-level restriction, while the downstream sLBP module still performs global point-level candidate selection.

\section{Experimental Results}
\label{sec:experiments}

\textbf{\textit{Dataset and implementation.}} We use ModelNet10 \citep{wu2015modelnet}, with 1024 normalized surface points per object. Following the controlled synthetic-deformation protocols used in previous non-rigid registration studies \citep{robustdefreg2025,deftransnet2025,synbench2024}, each source cloud is smoothly deformed at a controlled level $\ell\in\{0.1,0.2,\ldots,0.9\}$. In addition to the non-rigid deformation, the evaluated pairs contain relative rigid pose variation; the applied source--target rotation angle is retained and used only for the descriptive rotation-stratified analysis in Table~\ref{tab:rotation}. Exact point identity is preserved by the generator, but the target order is shuffled before it is given to the network. The standard 3,991-mesh training split is used for training; the 908 standard test meshes are split deterministically and by deformation level into 182 validation objects and 726 final test objects.

The point and patch objectives are optimized jointly. Patch-specific settings are those defined in Sec.~\ref{sec:method}: 32 patches, at most 24 retained members per patch for overlap supervision, 128-dimensional patch tokens, one four-head interaction block, overlap threshold 0.10, and residual strength 0.10. DefTransNet is the closest architectural reference because CoCo-Reg keeps its dense feature and sLBP backbone; Robust-DefReg provides a second recent non-rigid baseline. All methods are evaluated on exactly the same source--target pairs and target permutations. One final checkpoint per method is used, so reported standard deviations describe variation across test objects rather than across independent training runs.

\textbf{\textit{Evaluation metrics.}} Because the synthetic deformation preserves point identity, the primary metric is mean correspondence error (MCE). Let $\hat S=\{s_i+\hat d_i\}_{i=1}^{N}$ be the registered source and, for source point $i$, let $e_i=\|s_i+\hat d_i-t_i\|_2$. The per-object MCE is
\begin{equation}
E_{\mathrm{MCE}}=\frac{1}{N}\sum_{i=1}^{N} e_i.
\label{eq:mce}
\end{equation}
This measures distance to the known corresponding target point, rather than to the nearest target point. We also report the symmetric unsquared Chamfer distance \citep{fan2017pointset},
\begin{equation}
E_{\mathrm{CD}}=\frac{1}{2}\!\left(\frac{1}{N}\sum_{\hat s\in\hat S}\min_{t\in T}\|\hat s-t\|_2+\frac{1}{N}\sum_{t\in T}\min_{\hat s\in\hat S}\|t-\hat s\|_2\right),
\label{eq:cd}
\end{equation}
and the bidirectional 95th-percentile Hausdorff distance (HD95) \citep{huttenlocher1993hausdorff}. With $d(A,B)=\{\min_{b\in B}\|a-b\|_2:a\in A\}$,
\begin{equation}
E_{\mathrm{HD95}}=\max\!\left(Q_{0.95}[d(\hat S,T)],\;Q_{0.95}[d(T,\hat S)]\right).
\label{eq:hd95}
\end{equation}
MCE evaluates the known correspondence, whereas Chamfer and HD95 evaluate the geometry of the registered and target point sets without using identity. We additionally report the median and 95th percentile of per-object MCE. As a complementary severe-error statistic, object failure is defined as $E_{\mathrm{MCE}}>0.1$ and point failure as the within-object fraction with $e_i>0.1$. The threshold is used only for this diagnostic analysis and does not affect training or the continuous metrics.

\textbf{\textit{Statistical analysis.}} All comparisons with the closest baseline are paired on the same 726 test objects. For the overall scalar metrics we report the mean paired difference, a 20,000-resample paired bootstrap 95\% confidence interval, a two-sided Wilcoxon signed-rank test, and the paired rank-biserial effect size $r_{rb}$. Win-rate confidence intervals are exact binomial intervals. Object-level failure is analyzed with a 20,000-resample paired bootstrap interval for the failure-rate difference and an exact McNemar test. Deformation- and category-stratified mean-difference intervals use 10,000 paired bootstrap resamples; the nine deformation-level Wilcoxon tests are Holm-corrected. These analyses quantify consistency across the fixed paired test set; they do not estimate training-to-training uncertainty because only one checkpoint per method is available.

\textbf{\textit{Overall correspondence accuracy.}} Table~\ref{tab:overall} gives the complete primary and complementary metrics. CoCo-Reg has the lowest MCE, lower distributional tail, fewer object- and point-level failures, and lower Chamfer and HD95 than both evaluated reference models. Relative to DefTransNet, mean MCE is 72.6\% lower and CoCo-Reg is better on 670 of 726 paired objects.

\begin{table}[t]
\caption{Overall results on 726 held-out objects. Mean-valued quantities are reported as mean $\pm$ SD across test objects. Median and P95 are quantiles of the 726 per-object MCE values. ``Objects with MCE $>0.1$'' reports the number and percentage of test objects whose mean correspondence error exceeds 0.1. ``Points with error $>0.1$'' reports the mean within-object percentage of points whose identity-specific correspondence error exceeds 0.1. Lower is better for all metrics.}
\label{tab:overall}
\centering
\small
\resizebox{\linewidth}{!}{%
\begin{tabular}{lccccccc}
\toprule
Method & MCE & Median & P95 & Objects with MCE $>0.1$ & Points with error $>0.1$ (\%) & Chamfer & HD95\\
\midrule
Robust-DefReg & $0.1921\pm0.2001$ & 0.1188 & 0.6287 & 402/726 (55.4\%) & $48.4\pm35.6\%$ & $0.0178\pm0.0100$ & $0.1066\pm0.0608$ \\
DefTransNet & $0.1993\pm0.2165$ & 0.1180 & 0.6865 & 389/726 (53.6\%) & $47.3\pm35.9\%$ & $0.0170\pm0.0094$ & $0.1015\pm0.0575$ \\
\textbf{CoCo-Reg} & $\mathbf{0.0547\pm0.0693}$ & \textbf{0.0379} & \textbf{0.1529} & \textbf{90/726 (12.4\%)} & $\mathbf{17.3\pm21.1\%}$ & $\mathbf{0.0108\pm0.0061}$ & $\mathbf{0.0682\pm0.0381}$ \\
\bottomrule
\end{tabular}}
\end{table}

\textbf{\textit{Paired statistical evidence.}} Table~\ref{tab:pairedstats} uses the paired design directly. For every continuous metric, the bootstrap interval lies entirely below zero and the paired rank-biserial effect is large and positive when oriented in favor of CoCo-Reg. These tests quantify consistency and effect magnitude on the common test set; they do not remove the checkpoint-training limitation discussed later.

\begin{table}[t]
\caption{Paired CoCo-Reg vs. DefTransNet statistics. $\Delta$ is CoCo-Reg$-$DefTransNet; negative values favor CoCo-Reg. Mean-difference CIs are 20,000-resample paired bootstrap intervals; win-rate CIs are exact binomial intervals. $r_{rb}$ is oriented so positive values favor CoCo-Reg.}
\label{tab:pairedstats}
\centering
\small
\resizebox{\linewidth}{!}{%
\begin{tabular}{lcccc}
\toprule
Metric & Mean $\Delta$ [95\% CI] & $r_{rb}$ & CoCo-Reg win rate [95\% CI] & Wilcoxon $p$ \\
\midrule
MCE & $-0.1447\;[-0.1585,-0.1314]$ & 0.960 & 92.3\% [90.1, 94.1] & $2.89\times10^{-111}$ \\
Chamfer & $-0.00623\;[-0.00657,-0.00588]$ & 0.964 & 93.1\% [91.0, 94.8] & $3.17\times10^{-112}$ \\
HD95 & $-0.03328\;[-0.03569,-0.03096]$ & 0.934 & 89.5\% [87.1, 91.7] & $1.83\times10^{-105}$ \\
Point failure rate & $-0.2994\;[-0.3200,-0.2789]$ & 0.973 & 88.8\% [86.3, 91.0] & $1.10\times10^{-106}$ \\
\bottomrule
\end{tabular}}
\end{table}

\begin{figure*}[t]
    \centering
    \includegraphics[width=0.8\textwidth]{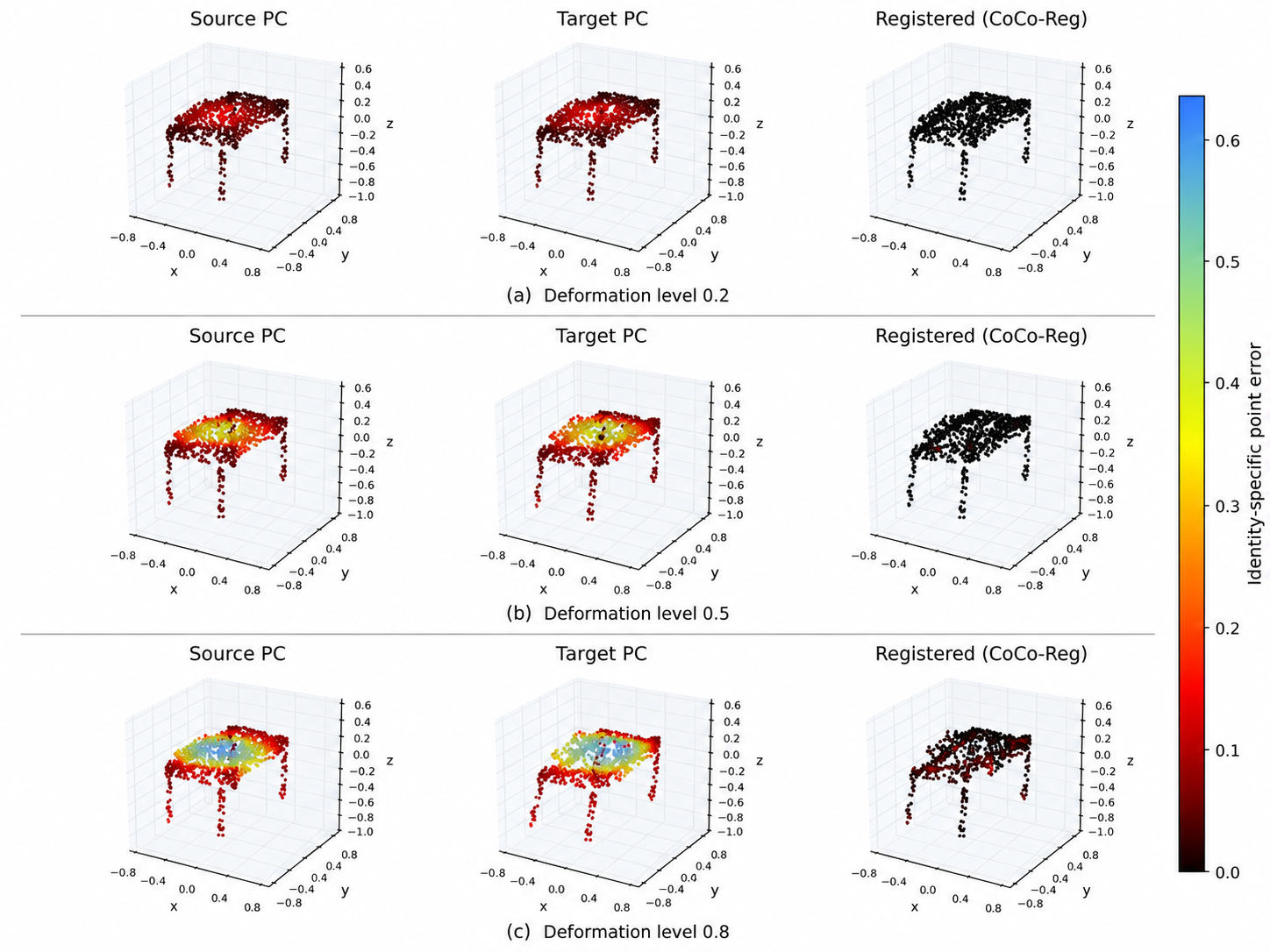}
    \caption{Qualitative registration under increasing deformation.
    Representative source, target, and CoCo-Reg registrations are shown at deformation levels 0.2, 0.5, and 0.8. Colors indicate the identity-specific correspondence error associated with each source-point identity and are propagated across the corresponding source, target, and registered points for spatial comparison.}
    \label{fig:qualitative}
\end{figure*}

\textbf{\textit{Robustness across deformation levels.}}
Table~\ref{tab:levels} shows the paired result across the complete tested deformation range. CoCo-Reg has lower MCE in every deformation stratum. Representative CoCo-Reg registrations at deformation levels 0.2, 0.5, and 0.8 are shown in Fig.~\ref{fig:qualitative}, providing a qualitative view of the predicted correspondence under increasing deformation. The paired bootstrap CI for CoCo-Reg$-$DefTransNet remains below zero at every level; all nine paired Wilcoxon tests remain significant after Holm correction, with the largest adjusted value $3.05\times10^{-9}$. The intermediate means are not interpreted as a pure deformation dose-response because the fixed evaluation also contains pose variation.

\begin{table}[t]
\caption{Deformation-stratified mean correspondence error (MCE). Means are mean $\pm$ SD across objects. $\Delta$ is the paired CoCo-Reg$-$DefTransNet mean difference with bootstrap 95\% CI; negative values favor CoCo-Reg.}
\label{tab:levels}
\centering
\scriptsize
\resizebox{\linewidth}{!}{%
\begin{tabular}{cccccccc}
\toprule
Level & $N$ & Robust-DefReg & DefTransNet & \textbf{CoCo-Reg} & $\Delta$ [95\% CI] & Win & Holm $p$ \\
\midrule
0.1 & 63 & $0.1111\pm0.1841$ & $0.1205\pm0.2055$ & $\mathbf{0.0293\pm0.0934}$ & $-0.0913[-0.1381,-0.0504]$ & 84.1\% & $3.05\times10^{-9}$ \\
0.2 & 80 & $0.1436\pm0.2104$ & $0.1485\pm0.2212$ & $\mathbf{0.0382\pm0.0967}$ & $-0.1103[-0.1530,-0.0713]$ & 85.0\% & $5.43\times10^{-11}$ \\
0.3 & 74 & $0.1346\pm0.2128$ & $0.1477\pm0.2366$ & $\mathbf{0.0336\pm0.0887}$ & $-0.1142[-0.1640,-0.0704]$ & 90.5\% & $5.43\times10^{-11}$ \\
0.4 & 92 & $0.1115\pm0.1221$ & $0.1167\pm0.1440$ & $\mathbf{0.0309\pm0.0288}$ & $-0.0858[-0.1156,-0.0600]$ & 94.6\% & $2.57\times10^{-14}$ \\
0.5 & 88 & $0.1676\pm0.1625$ & $0.1768\pm0.1798$ & $\mathbf{0.0438\pm0.0312}$ & $-0.1330[-0.1688,-0.1004]$ & 92.0\% & $1.13\times10^{-14}$ \\
0.6 & 80 & $0.2143\pm0.1870$ & $0.2245\pm0.2097$ & $\mathbf{0.0537\pm0.0341}$ & $-0.1708[-0.2152,-0.1292]$ & 92.5\% & $8.98\times10^{-14}$ \\
0.7 & 81 & $0.2591\pm0.2029$ & $0.2610\pm0.2224$ & $\mathbf{0.0720\pm0.0503}$ & $-0.1890[-0.2346,-0.1455]$ & 96.3\% & $4.34\times10^{-14}$ \\
0.8 & 82 & $0.2392\pm0.1833$ & $0.2367\pm0.1833$ & $\mathbf{0.0798\pm0.0542}$ & $-0.1569[-0.1923,-0.1246]$ & 92.7\% & $4.75\times10^{-14}$ \\
0.9 & 86 & $0.3285\pm0.2170$ & $0.3432\pm0.2385$ & $\mathbf{0.1039\pm0.0760}$ & $-0.2393[-0.2845,-0.1968]$ & 100.0\% & $7.19\times10^{-15}$ \\
\bottomrule
\end{tabular}}
\end{table}

\begin{table}[hpt]
\caption{Deformation-stratified set-level geometry. Values are mean $\pm$ SD across objects. CD is symmetric unsquared Chamfer distance and HD95 is the 95th-percentile bidirectional Hausdorff distance. Lower is better.}
\label{tab:setlevels}
\centering
\scriptsize
\resizebox{\linewidth}{!}{%
\begin{tabular}{c|ccc|ccc}
\toprule
& \multicolumn{3}{c|}{Chamfer distance (CD)} & \multicolumn{3}{c}{HD95} \\
Level & Robust & DefTransNet & \textbf{CoCo-Reg} & Robust & DefTransNet & \textbf{CoCo-Reg} \\
\midrule
0.1 & $0.0097\pm0.0087$ & $0.0093\pm0.0083$ & $\mathbf{0.0056\pm0.0046}$ & $0.0558\pm0.0485$ & $0.0534\pm0.0457$ & $\mathbf{0.0358\pm0.0293}$ \\
0.2 & $0.0114\pm0.0085$ & $0.0113\pm0.0086$ & $\mathbf{0.0066\pm0.0048}$ & $0.0670\pm0.0508$ & $0.0682\pm0.0541$ & $\mathbf{0.0422\pm0.0308}$ \\
0.3 & $0.0112\pm0.0084$ & $0.0108\pm0.0080$ & $\mathbf{0.0069\pm0.0049}$ & $0.0678\pm0.0494$ & $0.0652\pm0.0463$ & $\mathbf{0.0447\pm0.0313}$ \\
0.4 & $0.0146\pm0.0070$ & $0.0139\pm0.0067$ & $\mathbf{0.0084\pm0.0039}$ & $0.0863\pm0.0398$ & $0.0822\pm0.0380$ & $\mathbf{0.0547\pm0.0239}$ \\
0.5 & $0.0178\pm0.0077$ & $0.0170\pm0.0073$ & $\mathbf{0.0105\pm0.0044}$ & $0.1070\pm0.0440$ & $0.1019\pm0.0431$ & $\mathbf{0.0667\pm0.0260}$ \\
0.6 & $0.0204\pm0.0098$ & $0.0191\pm0.0086$ & $\mathbf{0.0119\pm0.0046}$ & $0.1225\pm0.0610$ & $0.1122\pm0.0526$ & $\mathbf{0.0731\pm0.0282}$ \\
0.7 & $0.0227\pm0.0085$ & $0.0211\pm0.0077$ & $\mathbf{0.0139\pm0.0048}$ & $0.1334\pm0.0529$ & $0.1254\pm0.0498$ & $\mathbf{0.0867\pm0.0297}$ \\
0.8 & $0.0234\pm0.0089$ & $0.0223\pm0.0083$ & $\mathbf{0.0146\pm0.0053}$ & $0.1411\pm0.0559$ & $0.1344\pm0.0529$ & $\mathbf{0.0931\pm0.0368}$ \\
0.9 & $0.0266\pm0.0072$ & $0.0257\pm0.0068$ & $\mathbf{0.0170\pm0.0057}$ & $0.1620\pm0.0478$ & $0.1558\pm0.0464$ & $\mathbf{0.1070\pm0.0360}$ \\
\bottomrule
\end{tabular}}
\end{table}

\begin{table}[hpt]
\caption{Rotation-stratified descriptive analysis of the final test set. Values are MCE $\pm$ SD across objects. The last group contains observed rotations above $60^\circ$ up to $91.7^\circ$.}
\label{tab:rotation}
\centering
\small
\begin{tabular}{cccccc}
\toprule
Rotation & $N$ & Robust-DefReg & DefTransNet & \textbf{CoCo-Reg} & Win vs. Def. \\
\midrule
$0$--$30^{\circ}$ & 234 & $0.0792\pm0.1018$ & $0.0771\pm0.1030$ & $\mathbf{0.0354\pm0.0465}$ & 80.3\% \\
$>30$--$60^{\circ}$ & 240 & $0.1160\pm0.1101$ & $0.1127\pm0.1112$ & $\mathbf{0.0487\pm0.0547}$ & 96.7\% \\
$>60^{\circ}$ & 252 & $0.3693\pm0.2134$ & $0.3954\pm0.2324$ & $\mathbf{0.0783\pm0.0898}$ & 99.2\% \\
\bottomrule
\end{tabular}
\end{table}

\begin{table}[!t]
\caption{Category-stratified CoCo-Reg vs. DefTransNet MCE. Mean $\pm$ SD are across objects; $\Delta$ is CoCo-Reg$-$DefTransNet with bootstrap 95\% CI. All ten categories are shown.}
\label{tab:category}
\centering
\scriptsize
\begin{minipage}[t]{0.495\linewidth}
\centering
\resizebox{\linewidth}{!}{%
\begin{tabular}{lccc}
\toprule
Category & DefTransNet & \textbf{CoCo-Reg} & $\Delta$ [95\% CI] / Win \\
\midrule
Bathtub & $0.2807\pm0.2607$ & $\mathbf{0.0863\pm0.0947}$ & $-0.1943[-0.2628,-0.1327]$ / 92.9\% \\
Bed & $0.2334\pm0.2806$ & $\mathbf{0.0518\pm0.0666}$ & $-0.1816[-0.2415,-0.1257]$ / 93.7\% \\
Chair & $0.1785\pm0.1720$ & $\mathbf{0.0457\pm0.0381}$ & $-0.1328[-0.1656,-0.1021]$ / 90.1\% \\
Desk & $0.2502\pm0.2476$ & $\mathbf{0.0620\pm0.0578}$ & $-0.1881[-0.2466,-0.1364]$ / 95.4\% \\
Dresser & $0.2048\pm0.1943$ & $\mathbf{0.0677\pm0.0951}$ & $-0.1371[-0.1784,-0.1002]$ / 94.1\% \\
\bottomrule
\end{tabular}}
\end{minipage}\hfill
\begin{minipage}[t]{0.495\linewidth}
\centering
\resizebox{\linewidth}{!}{%
\begin{tabular}{lccc}
\toprule
Category & DefTransNet & \textbf{CoCo-Reg} & $\Delta$ [95\% CI] / Win \\
\midrule
Monitor & $0.1349\pm0.1330$ & $\mathbf{0.0473\pm0.0448}$ & $-0.0876[-0.1099,-0.0681]$ / 92.6\% \\
Night stand & $0.2124\pm0.2382$ & $\mathbf{0.0675\pm0.1041}$ & $-0.1449[-0.1939,-0.0992]$ / 85.1\% \\
Sofa & $0.1989\pm0.2240$ & $\mathbf{0.0415\pm0.0464}$ & $-0.1573[-0.2008,-0.1179]$ / 96.5\% \\
Table & $0.2228\pm0.2352$ & $\mathbf{0.0531\pm0.0828}$ & $-0.1697[-0.2192,-0.1240]$ / 92.2\% \\
Toilet & $0.1313\pm0.1279$ & $\mathbf{0.0453\pm0.0436}$ & $-0.0860[-0.1068,-0.0664]$ / 90.1\% \\
\bottomrule
\end{tabular}}
\end{minipage}
\end{table}

\textbf{\textit{Set-level geometry across deformation.}} The correspondence result is not specific to MCE. Table~\ref{tab:setlevels} reports Chamfer and HD95 at the same nine deformation levels. CoCo-Reg is lower than both reference checkpoints for both set-level metrics in every stratum. Thus, the deformation-stratified advantage is visible both when ground-truth identity is enforced by MCE and when registration quality is evaluated only through nearest-neighbor geometry.

\textbf{\textit{Severe correspondence failures.}} The reduction in failure rate is also strongly paired. DefTransNet fails on 389/726 objects and CoCo-Reg on 90/726, an absolute reduction of 41.18 percentage points with paired bootstrap 95\% CI [37.47, 44.90]. Among the 305 discordant pairs, 302 change from DefTransNet failure to CoCo-Reg success and only 3 change in the opposite direction; exact McNemar $p=1.45\times10^{-85}$. The within-object point-failure result in Tables~\ref{tab:overall}--\ref{tab:pairedstats} shows the same pattern at finer resolution.

\textbf{\textit{Additional rotation-stratified robustness.}} Rotation is not a primary claim of this study, but the same 726 evaluated outputs can be grouped descriptively without new inference. Table~\ref{tab:rotation} shows that CoCo-Reg remains lower in all three observed rotation groups. We do not interpret this as a controlled claim of rotation invariance or extrapolation because the available checkpoints were not produced as a matched rotation-specific study.

\textbf{\textit{Category consistency.}} The aggregate advantage is not dominated by one ModelNet10 class. Table~\ref{tab:category} shows a negative CoCo-Reg$-$DefTransNet paired difference in all ten categories; every category bootstrap interval remains below zero and category-wise paired win rates range from 85.1\% to 96.5\%.

\section{Discussion and Outlook}

The results support the central design of CoCo-Reg: dense non-rigid correspondence benefits from regional geometric information, but this information does not need to be converted into a hard restriction on the final point-level search. By projecting patch-level information back to the dense descriptors, CoCo-Reg preserves the flexibility of global point matching while providing each point with a larger spatial context. Across the fixed paired evaluation, this leads to substantially lower correspondence error than the evaluated reference methods, and the same trend is observed across all tested deformation levels. The improvements in Chamfer distance and HD95 further show that the gain is not limited to identity-specific correspondence error, but is also reflected in the geometry of the registered point set.

A second contribution is the patch-supervision strategy. Because source and target patches are constructed independently, a source patch may overlap with several valid target patches after deformation. Treating patch correspondence as a single-label classification problem would therefore impose an unnecessarily rigid target. The identity-corrected, multi-positive formulation instead allows several geometrically consistent patch relations to contribute to supervision while still encouraging the strongest overlap through the complementary best-overlap objective. This makes the patch branch better aligned with the non-rigid setting, where regional boundaries need not remain identical after deformation.

The paired analysis also shows that the improvement is systematic rather than being driven by a small number of favorable cases. CoCo-Reg performs better on the large majority of paired test objects, reduces the upper tail of the correspondence-error distribution, and markedly lowers both object-level and point-level severe-error rates. The same direction is observed across deformation strata and object categories. These results are important for dense deformation estimation because a small number of large correspondence errors can lead to locally incorrect displacement fields even when an average set-distance metric remains acceptable.

The present study evaluates the complete CoCo-Reg configuration on one final checkpoint per method and on controlled synthetic deformations of complete ModelNet10 shapes. The main next step is therefore to test whether the same advantage persists under more realistic conditions, including partial overlap, noise, outliers, topology changes, and real deforming surfaces such as soft tissue. A matched multi-seed study and separately trained hard-routing variants would also make it possible to isolate more precisely which parts of the patch-context design contribute most strongly to the observed gain. Beyond this, adaptive patch construction and confidence-weighted regional information could further improve the method when the reliability of patch context varies across the shape.

\section{Conclusion}
We presented CoCo-Reg, a dense non-rigid point-cloud registration method that introduces regional patch information without using patch predictions to restrict the final point-level correspondence search. The method combines independently constructed geometric patches, source--target patch interaction, identity-corrected multi-positive patch supervision, and residual projection of regional information back to dense point features. On 726 paired held-out ModelNet10 objects across deformation levels 0.1--0.9, CoCo-Reg consistently reduces correspondence error compared with the evaluated reference methods. The improvement is also reflected in Chamfer distance, HD95, tail-error statistics, and severe correspondence failures, indicating that the benefit extends beyond a reduction in average error.
Overall, the results support the use of regional context as guidance for dense correspondence while preserving a globally searchable point-level matching stage. Future work will examine the same principle under partial, noisy, and real deforming observations and will further isolate the contribution of individual patch-context components through matched multi-seed and hard-routing comparisons.

\bibliographystyle{plainnat}
\bibliography{references}

@article{monjiazad2023review,
  author  = {Monji-Azad, Sara and Hesser, J{\"u}rgen and L{\"o}w, Nikolas},
  title   = {A Review of Non-Rigid Transformations and Learning-Based 3D Point Cloud Registration Methods},
  journal = {ISPRS Journal of Photogrammetry and Remote Sensing},
  volume  = {196},
  pages   = {58--72},
  year    = {2023},
  doi     = {10.1016/j.isprsjprs.2022.12.023}
}

@article{maennle2023digitaltwin,
  author  = {M{\"a}nnle, David and Pohlmann, Jan and Monji-Azad, Sara and Hesser, J{\"u}rgen and Rotter, Nicole and Affolter, Annette and Lammert, Anne and Kramer, Benedikt and Ludwig, Sonja and Huber, Lena and Scherl, Claudia},
  title   = {Artificial Intelligence Directed Development of a Digital Twin to Measure Soft Tissue Shift During Head and Neck Surgery},
  journal = {PLOS ONE},
  volume  = {18},
  number  = {8},
  pages   = {e0287081},
  year    = {2023},
  doi     = {10.1371/journal.pone.0287081}
}

@article{monjiazad2024digitaltwin,
  author  = {Monji-Azad, Sara and M{\"a}nnle, David and Hesser, J{\"u}rgen and Pohlmann, Jan and Rotter, Nicole and Affolter, Annette and Weis, Cleo Aron and Ludwig, Sonja and Scherl, Claudia},
  title   = {Point Cloud Registration for Measuring Shape Dependence of Soft Tissue Deformation by Digital Twins in Head and Neck Surgery},
  journal = {Biomedicine Hub},
  volume  = {9},
  number  = {1},
  pages   = {9--15},
  year    = {2024},
  doi     = {10.1159/000535421}
}

@article{robustdefreg2025,
  author  = {Monji-Azad, Sara and Kinz, Marvin and M{\"a}nnel, David and Scherl, Claudia and Hesser, J{\"u}rgen},
  title   = {Robust-DefReg: A Robust Coarse to Fine Non-Rigid Point Cloud Registration Method Based on Graph Convolutional Neural Networks},
  journal = {Measurement Science and Technology},
  volume  = {36},
  number  = {1},
  pages   = {015426},
  year    = {2025},
  doi     = {10.1088/1361-6501/ad916c}
}

@article{deftransnet2025,
  author  = {Monji-Azad, Sara and Kinz, Marvin and Kothari, Siddharth and Khanna, Robin and Mihan, Amrei Carla and M{\"a}nnel, David and Scherl, Claudia and Hesser, J{\"u}rgen},
  title   = {DefTransNet: A Transformer-Based Method for Non-Rigid Point Cloud Registration in the Simulation of Soft Tissue Deformation},
  journal = {Measurement Science and Technology},
  volume  = {36},
  number  = {7},
  pages   = {076006},
  year    = {2025},
  doi     = {10.1088/1361-6501/ade613}
}

@inproceedings{wang2019dcp,
  author    = {Wang, Yue and Solomon, Justin M.},
  title     = {Deep Closest Point: Learning Representations for Point Cloud Registration},
  booktitle = {Proceedings of the IEEE/CVF International Conference on Computer Vision (ICCV)},
  pages     = {3523--3532},
  year      = {2019}
}

@inproceedings{yew2020rpmnet,
  author    = {Yew, Zi Jian and Lee, Gim Hee},
  title     = {RPM-Net: Robust Point Matching Using Learned Features},
  booktitle = {Proceedings of the IEEE/CVF Conference on Computer Vision and Pattern Recognition (CVPR)},
  pages     = {11824--11833},
  year      = {2020}
}

@inproceedings{huang2021predator,
  author    = {Huang, Shengyu and Gojcic, Zan and Usvyatsov, Mikhail and Wieser, Andreas and Schindler, Konrad},
  title     = {PREDATOR: Registration of 3D Point Clouds with Low Overlap},
  booktitle = {Proceedings of the IEEE/CVF Conference on Computer Vision and Pattern Recognition (CVPR)},
  pages     = {4267--4276},
  year      = {2021}
}

@inproceedings{bai2021pointdsc,
  author    = {Bai, Xuyang and Luo, Zixin and Zhou, Lei and Chen, Hongkai and Li, Lei and Hu, Zeyu and Fu, Hongbo and Tai, Chiew-Lan},
  title     = {PointDSC: Robust Point Cloud Registration Using Deep Spatial Consistency},
  booktitle = {Proceedings of the IEEE/CVF Conference on Computer Vision and Pattern Recognition (CVPR)},
  pages     = {15859--15869},
  year      = {2021},
  doi       = {10.1109/CVPR46437.2021.01560}
}

@inproceedings{yew2022regtr,
  author    = {Yew, Zi Jian and Lee, Gim Hee},
  title     = {REGTR: End-to-End Point Cloud Correspondences with Transformers},
  booktitle = {Proceedings of the IEEE/CVF Conference on Computer Vision and Pattern Recognition (CVPR)},
  pages     = {6677--6686},
  year      = {2022},
  doi       = {10.1109/CVPR52688.2022.00656}
}

@inproceedings{puy2020flot,
  author    = {Puy, Gilles and Boulch, Alexandre and Marlet, Renaud},
  title     = {FLOT: Scene Flow on Point Clouds Guided by Optimal Transport},
  booktitle = {Computer Vision -- ECCV 2020},
  pages     = {527--544},
  year      = {2020}
}

@inproceedings{li2022lepard,
  author    = {Li, Yang and Harada, Tatsuya},
  title     = {Lepard: Learning Partial Point Cloud Matching in Rigid and Deformable Scenes},
  booktitle = {Proceedings of the IEEE/CVF Conference on Computer Vision and Pattern Recognition (CVPR)},
  pages     = {5554--5564},
  year      = {2022}
}

@inproceedings{li2022ndp,
  author    = {Li, Yang and Harada, Tatsuya},
  title     = {Non-Rigid Point Cloud Registration with Neural Deformation Pyramid},
  booktitle = {Advances in Neural Information Processing Systems},
  volume    = {35},
  pages     = {27757--27768},
  year      = {2022}
}

@article{dine2026,
  author  = {Monji-Azad, Sara and Beer, Rohit and Kinz, Marvin and Scherl, Claudia and Hesser, J{\"u}rgen},
  title   = {DINE: Distance Is Not Enough---Learning Global Deformation Priors for Robust Soft-Tissue Point Cloud Registration},
  journal = {arXiv preprint arXiv:2607.14946},
  year    = {2026},
  doi     = {10.48550/arXiv.2607.14946}
}

@inproceedings{qi2017pointnetpp,
  author    = {Qi, Charles Ruizhongtai and Yi, Li and Su, Hao and Guibas, Leonidas J.},
  title     = {PointNet++: Deep Hierarchical Feature Learning on Point Sets in a Metric Space},
  booktitle = {Advances in Neural Information Processing Systems},
  volume    = {30},
  pages     = {5099--5108},
  year      = {2017}
}

@article{wang2019dgcnn,
  author  = {Wang, Yue and Sun, Yongbin and Liu, Ziwei and Sarma, Sanjay E. and Bronstein, Michael M. and Solomon, Justin M.},
  title   = {Dynamic Graph CNN for Learning on Point Clouds},
  journal = {ACM Transactions on Graphics},
  volume  = {38},
  number  = {5},
  pages   = {146:1--146:12},
  year    = {2019},
  doi     = {10.1145/3326362}
}

@inproceedings{yu2021cofinet,
  author    = {Yu, Hao and Li, Fu and Saleh, Mahdi and Busam, Benjamin and Ilic, Slobodan},
  title     = {CoFiNet: Reliable Coarse-to-Fine Correspondences for Robust Point Cloud Registration},
  booktitle = {Advances in Neural Information Processing Systems},
  volume    = {34},
  pages     = {23872--23884},
  year      = {2021}
}

@article{qin2023geotransformer,
  author  = {Qin, Zheng and Yu, Hao and Wang, Changjian and Guo, Yulan and Peng, Yuxing and Ilic, Slobodan and Hu, Dewen and Xu, Kai},
  title   = {GeoTransformer: Fast and Robust Point Cloud Registration with Geometric Transformer},
  journal = {IEEE Transactions on Pattern Analysis and Machine Intelligence},
  volume  = {45},
  number  = {8},
  pages   = {9806--9821},
  year    = {2023},
  doi     = {10.1109/TPAMI.2023.3259038}
}

@inproceedings{yu2022pointbert,
  author    = {Yu, Xumin and Tang, Lulu and Rao, Yongming and Huang, Tiejun and Zhou, Jie and Lu, Jiwen},
  title     = {Point-BERT: Pre-Training 3D Point Cloud Transformers with Masked Point Modeling},
  booktitle = {Proceedings of the IEEE/CVF Conference on Computer Vision and Pattern Recognition (CVPR)},
  pages     = {19313--19322},
  year      = {2022}
}

@inproceedings{pang2022pointmae,
  author    = {Pang, Yatian and Wang, Wenxiao and Tay, Francis E. H. and Liu, Wei and Tian, Yonghong and Yuan, Li},
  title     = {Masked Autoencoders for Point Cloud Self-Supervised Learning},
  booktitle = {Computer Vision -- ECCV 2022},
  pages     = {604--621},
  year      = {2022}
}

@inproceedings{chen2023pointgpt,
  author    = {Chen, Guangyan and Wang, Meiling and Yang, Yi and Yu, Kai and Yuan, Li and Yue, Yufeng},
  title     = {PointGPT: Auto-Regressively Generative Pre-Training from Point Clouds},
  booktitle = {Advances in Neural Information Processing Systems},
  volume    = {36},
  year      = {2023}
}

@inproceedings{wu2015modelnet,
  author    = {Wu, Zhirong and Song, Shuran and Khosla, Aditya and Yu, Fisher and Zhang, Linguang and Tang, Xiaoou and Xiao, Jianxiong},
  title     = {3D ShapeNets: A Deep Representation for Volumetric Shapes},
  booktitle = {Proceedings of the IEEE Conference on Computer Vision and Pattern Recognition (CVPR)},
  pages     = {1912--1920},
  year      = {2015}
}

@inproceedings{fan2017pointset,
  author    = {Fan, Haoqiang and Su, Hao and Guibas, Leonidas J.},
  title     = {A Point Set Generation Network for 3D Object Reconstruction from a Single Image},
  booktitle = {Proceedings of the IEEE Conference on Computer Vision and Pattern Recognition (CVPR)},
  pages     = {605--613},
  year      = {2017},
  doi       = {10.1109/CVPR.2017.264}
}

@article{huttenlocher1993hausdorff,
  author  = {Huttenlocher, Daniel P. and Klanderman, Gregory A. and Rucklidge, William J.},
  title   = {Comparing Images Using the Hausdorff Distance},
  journal = {IEEE Transactions on Pattern Analysis and Machine Intelligence},
  volume  = {15},
  number  = {9},
  pages   = {850--863},
  year    = {1993},
  doi     = {10.1109/34.232073}
}

@article{synbench2024,
  author  = {Monji-Azad, Sara and Kinz, Marvin and Scherl, Claudia and M{\"a}nnle, David and Hesser, J{\"u}rgen and L{\"o}w, Nikolas},
  title   = {SynBench: A Synthetic Benchmark for Non-Rigid 3D Point Cloud Registration},
  journal = {arXiv preprint arXiv:2409.14474},
  year    = {2024},
  doi     = {10.48550/arXiv.2409.14474}
}

\end{document}